\pdfoutput=1
\documentclass[11pt]{article}

\usepackage[]{ACL2023}

\usepackage{times}
\usepackage{latexsym}
\usepackage{multirow}
\usepackage{pgfplots}
\usepackage{pgfplotstable}
\usepackage{filecontents}
\usepackage{amsmath}
\usepackage{colortbl}
\usepackage{xcolor}
\usepackage{booktabs}
\usepackage{tcolorbox}
\usepackage{epigraph}

\usepackage[T1]{fontenc}
\usepackage[utf8]{inputenc}

\usepackage{microtype}
\usepgfplotslibrary{colorbrewer}
\pgfplotsset{compat = 1.15, cycle list/Set1-8} 
\usetikzlibrary{pgfplots.statistics, pgfplots.colorbrewer} 
\usepackage{pgfplotstable}
\usepackage{filecontents}
\usepackage{varwidth}
\usepackage{soul}
\usepackage[export]{adjustbox}
\usepackage{dblfloatfix}

\usetikzlibrary{pgfplots.groupplots}
\usetikzlibrary{shapes.geometric}
\usetikzlibrary{positioning,fit,shapes.geometric,backgrounds, calc}
\usetikzlibrary{arrows,decorations.markings}
\usetikzlibrary{shapes.arrows}
\usetikzlibrary{patterns}
\tikzset{textnode/.style={inner sep=0pt,outer sep=0,execute at begin node={\strut}}}
\tikzstyle{state} = [textnode,circle, draw, inner sep=0pt, outer sep=0]
\usepgfplotslibrary{groupplots}

\definecolor{ppt-purple}{RGB}{112, 48, 160}
\definecolor{ppt-blue}{RGB}{91, 155, 213}
\definecolor{ppt-orange}{RGB}{237, 125, 49}
\definecolor{ppt-gray}{RGB}{165, 165, 165}
\definecolor{ppt-yellow}{RGB}{255, 192, 0}
\definecolor{ppt-darkblue}{RGB}{68, 114, 196}
\definecolor{ppt-green}{RGB}{112, 173, 71}
\definecolor{ppt-red}{RGB}{255, 0, 0}

\pgfplotsset{every axis/.append style={
    xlabel={$x$},          % default put x on x-axis
    ylabel={$y$},          % default put y on y-axis
    label style={font=\sffamily},
    tick label style={font=\sffamily\scriptsize},
    xticklabel style = {font=\sffamily\scriptsize},
    title style = {font=\normalsize\sffamily},
    ylabel near ticks,
    y label style={font=\sffamily\small},
    xlabel near ticks,
    x label style={font=\sffamily\small},
    legend cell align={left},
    legend style={draw=none, font=\sffamily\scriptsize},
    },
    legend image code/.code={
    \draw[mark repeat=2,mark phase=2]
        plot coordinates {
        (0cm,0cm)
        (0.15cm,0cm)        %% default is (0.3cm,0cm)
        (0.3cm,0cm)         %% default is (0.6cm,0cm)
        };%
    }
    }
\pgfplotsset{compat=1.17}

\usepackage{inconsolata}

\newcommand*{\eg}{{\em e.g.}}

\newcommand*{\ie}{{\em i.e.}}

\title{The Power of Framing: How News Headlines Guide Search Behavior}

\author{
  Amrit Poudel\textsuperscript{1} \quad 
  Maria Milkowski \textsuperscript{1,2} \quad 
  Tim Weninger\textsuperscript{1,2} \\
  \textsuperscript{1}Department of Computer Science and Engineering, \\
  \textsuperscript{2}ND-IBM Technology Ethics Lab, \\
  University of Notre Dame, Notre Dame, IN, USA \\
  \texttt{\{apoudel, mmilkows, tweninger\}@nd.edu} \\
}

\begin{document}

\maketitle

\begin{abstract}

Search engines play a central role in how people gather information, but subtle cues like headline \textit{framing} may influence not only what users believe but also how they search. While framing effects on judgment are well documented, their impact on subsequent search behavior is less understood. We conducted a controlled experiment where participants issued queries and selected from headlines filtered by specific linguistic frames. Headline framing significantly shaped follow-up queries: conflict and strategy frames disrupted alignment with prior selections, while episodic frames led to more concrete queries than thematic ones. We also observed modest short-term frame persistence that declined over time. These results suggest that even brief exposure to framing can meaningfully alter the direction of users’ information-seeking behavior. %All data and analysis scripts are available at \url{https://doi.org/10.7910/DVN/1MUI68}.

% These results demonstrate the enduring impact of media framing on online information-seeking and have implications for search engine design, digital literacy, and public discourse in algorithmically curated environments.

\end{abstract}

\section{Introduction}

\epigraph{
``You can’t see or hear frames. They are part of what we cognitive scientists call the “cognitive unconscious”—structures in our brains that we cannot consciously access, but know by their consequences.''
}{
~\cite{lakoff2014all}
}

About two-thirds of U.S. adults get news at least occasionally from websites or apps (68\%) and from search engines (65\%)~\cite{pew2021digitalnews}. This shift from traditional media to digital platforms has transformed how people access and interpret information. Search engines and social media now act as gatekeepers, shaping public perception by controlling which stories appear most prominently~\cite{goldman2005search, introna2000shaping, poudel2024navigating, poudel2025digital}.

\begin{figure}[t]
    \centering
    \includegraphics[width=0.99\linewidth]{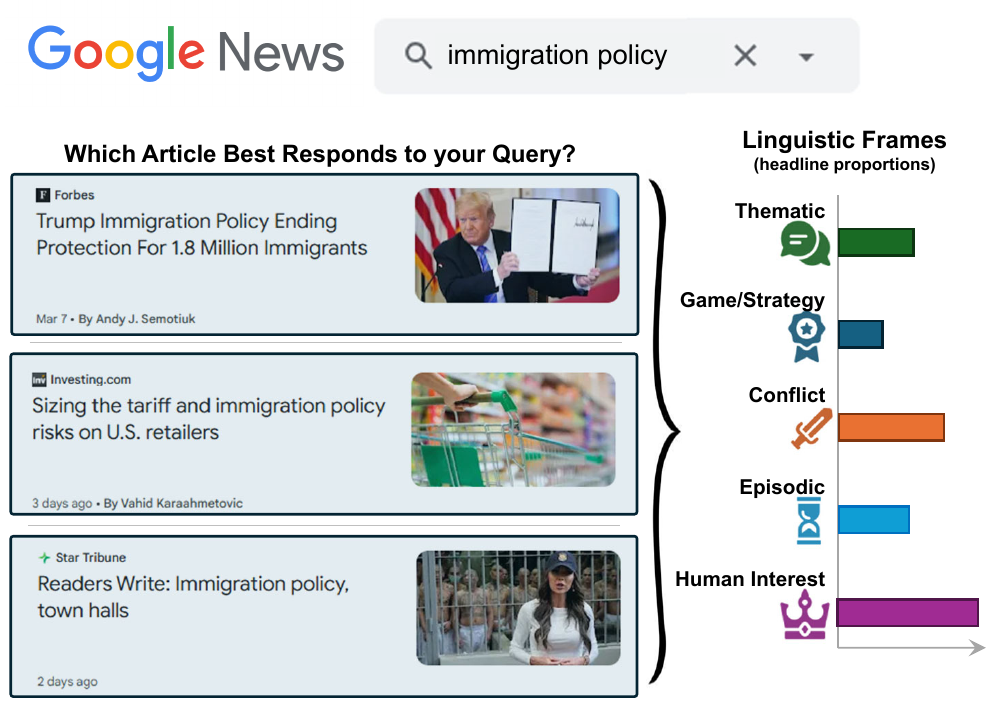}
    \caption{News headlines have linguistic frames. Here we ask: do these headline frames present in search results change search behavior?}
    \label{fig:features}
    \vspace{-.2cm}
\end{figure}

While the visibility and ranking of news content receive significant attention~\cite{diakopoulos2015algorithmic, lazer2018science, robertson2018auditing}, far less is known about how \textit{framing} of headlines within those results influences behavior. Framing, whether through linguistic cues, emphasis, or implied meaning, can shape perceptions, reinforce bias, and steer the course of information seeking~\cite{guo2025news}. Algorithmic ranking amplifies certain narratives~\cite{noble2018algorithms}, and exposure to specific frames affects how issues are interpreted~\cite{lecheler2012news, liu2019detecting, rathje2017power}. Yet how framing within results affects subsequent search behavior remains poorly understood.

Discussions of search engine bias~\cite{mowshowitz2002assessing} and query reformulation~\cite{boldi2011query} often center around how users revise their queries based on internal goals. But search behavior may also be influenced in other ways like the framing of content in top results~\cite{chen2018cognitive} as shown in Fig.~\ref{fig:features}. Click patterns and term overlap are often used to track engagement~\cite{wang2008mining}, though this overlooks more subtle semantic shifts that reflect deeper interpretive changes.

Because headlines frequently serve as the first point of contact with a topic~\cite{konnikova2014headlines}, these frames may exert a lasting influence on how people explore, understand, and act upon information~\cite{lecheler2011getting}.

% \paragraph{Framing Theory} 
\subsection*{Framing and Cognitive Mechanisms of Influence}

Framing, a concept rooted in psychology~\cite{goffman1974frame}, refers to how information is structured to emphasize certain aspects of reality, shaping interpretation, evaluation, and response~\cite{entman1993framing}. In the context of news communication, framing can define problems, diagnose causes, make moral judgments, and suggest solutions~\cite{de2005news}. Common cognitive frames (\eg, conflict, human interest, and strategy) shape how issues are perceived and discussed~\cite{neuman1992common, semetko2000framing}. While these patterns are well documented in traditional media~\cite{gans1979deciding, tuchman1978making, cooper2002media}, much less is known about how framing operates within digital platforms and how it influences beliefs and behaviors in interactive settings.

Frames shape how people process information by activating cognitive biases that guide interpretation and decision-making~\cite{tversky1981framing}. Three such mechanisms are especially relevant in digital environments: 

\noindent\textbf{Recency.} People are more likely to recall and act on information they encountered most recently~\cite{tversky1974judgment}. Frames that dominate the news cycle become disproportionately salient in memory and judgment. 

\noindent\textbf{Availability.} Individuals rely on easily accessible information when forming judgments~\cite{tversky1974judgment, kahneman2011thinking}. Repeated exposure to specific frames can cause them to feel representative, even when alternative perspectives exist. 

\noindent\textbf{Priming.} Early exposure to particular language or narratives shapes how users interpret subsequent information~\cite{bargh1996automaticity}. For example, describing an event as a ``crisis'' rather than a ``dispute'' primes users to expect urgency and instability.

Through digital media’s rapid and repetitive cycles, these effects are amplified. Framing becomes more than a feature of individual stories—it shapes the broader narrative environment in which public discourse unfolds.

\subsection*{Do Headline Frames Affect Search Behavior?}

The present work examines how the framing of search engine results influences user behavior, particularly in shaping subsequent queries and altering the trajectory of information seeking. Drawing on framing theory and cognitive mechanisms, we investigate how repeated exposure to framed content may guide what users search for next, how they interpret what they encounter, and how their mental models evolve over time. Specifically, we formulate the following hypotheses:

\begin{table*}[t]
\centering
\caption{Framing Categories in News Headlines}
\vspace{-.4cm}
\footnotesize{
\begin{tabular}{@{}l |l l@{}}
\toprule
\textbf{Frame} & \textbf{Description} & \textbf{\# Headlines} \\
\midrule
\textbf{Game/Strategy} & Emphasizes tactics, strategy, or competition in political or institutional contexts. & 42,171 \\
\textbf{Conflict} & Highlights polarization, disputes, or antagonism between individuals or groups. & 28,367 \\
\textbf{Thematic} & Focuses on systemic issues, policies, or trends with broader social relevance. & 56,627 \\
\textbf{Episodic} & Focuses on specific events, incidents, or individuals without broader context. & 10,698 \\
\textbf{Human Interest} & Centers on emotional appeals or personal stories that elicit empathy or drama. & 16,593 \\
\textbf{Other} & Does not clearly fit into the above categories or lacks identifiable framing. & 3,371 \\
\bottomrule
\end{tabular}
}
\label{tab:framing_categories}
\end{table*}

\begin{description}
    \item[\textbf{H1}] \textbf{Framing Influences Search Behavior.}
    \textit{(Priming \& Availability)} Framing will shape follow-up queries by activating certain concepts and interpretations.
    \item[\textbf{H2}] \textbf{Frame Exposure Leads to Short-Term Persistence.} 
    \textit{(Priming \& Availability)} Exposure to a particular narrative frame increases the likelihood that subsequent queries retrieve similarly framed headlines.%, as initial framing primes users to reinforce the same interpretive structure.
    \item[\textbf{H3}] \textbf{Framing Effects Accumulate Over Time.} 
    \textit{(Recency)} If framing effects persist, users will increasingly retrieve headlines matching the initial frame in later rounds.
\end{description}

We also explore how individual differences may moderate framing effects:

\begin{description}
    \item[\textbf{E1}] \textbf{Framing Effects Interact with Demographics.} \textit{(Cognitive Availability)} Frequent news seekers may show stronger frame persistence across queries.

    \item[\textbf{E2}] \textbf{Political Orientation Influences Frame Selection.}  \textit{(Motivated Reasoning)} Participants with stronger ideological views may respond more to frames that align with their beliefs.
\end{description}

\paragraph{Findings in Brief} 
Our results show that framing meaningfully shapes the trajectory of user queries (supporting H1). Exposure to conflict and game/strategy frames reduces semantic alignment with clicked headlines, suggesting a shift toward emotionally or ideologically salient interpretations. We also observe short-term frame persistence: users tend to submit follow-up queries that retrieve similarly framed content (supporting H2), though this effect does not compound over time (no support for H3).

Exploratory findings suggest that frequent news seekers and politically liberal users may be more sensitive to framing effects (tentative support for E1 and E2). Overall, these results reveal subtle but consistent ways headline framing can influence search behavior after minimal exposure. 
All data and analysis scripts are available at \url{https://doi.org/10.7910/DVN/1MUI68}.

\section{Related Work and Background}

Framing plays a central role in shaping how individuals interpret and act on information in digital environments. Prior work has established that: (1) headlines serve as powerful framing devices, shaping perception through linguistic and structural cues; (2) these frames influence user behavior by engaging cognitive biases that guide attention, memory, and decision-making; and (3) framing can be systematically analyzed in search contexts using established taxonomies from political communication and media studies.

\subsection{Headlines as Framing Devices}

Headlines are powerful framing tools, often the first (and sometimes only) text users read when scanning the news~\cite{konnikova2014headlines}. Readers act as ``shoppers of headlines,'' forming impressions based on brief linguistic cues~\cite{english1944study}. Headlines do more than summarize; they signal salience, shape interpretation, and prime cognitive associations~\cite{papacharissi2018importance}.

This effect is well documented. During the 2009 H1N1 outbreak, headlines emphasized risk, conflict, and strategy over factual content~\cite{kee2010framing}. More recent computational work shows that structural features in headlines can even guide algorithmic inference. For example, stock movements have been predicted using only the emotional framing of financial news headlines~\cite{bhat2024stock}. 

\subsection{Framing and User Behavior}

Headline framing engages core cognitive mechanisms that shape how people process, recall, and act on information. Psychological theories emphasize the role of heuristics \eg, availability, recency, and priming, when navigating online information environments.

Emotionally charged or dramatized frames can drive offline behavior~\cite{brader2005striking, gross2004framing}. During political unrest, vivid online narratives increase protest participation even when internet access itself does not explain mobilization~\cite{ruijgrok2017web}. In search environments, Subtle phrasing differences can shift beliefs without awareness---a phenomenon known as the Search Engine Manipulation Effect (SEME)~\cite{epstein2015search}. Public health research shows similar patterns: fentanyl-related queries closely track overdose deaths~\cite{arendt2021opioid}.

Across these domains, framing activates interpretive frameworks that steer attention, guide further search, and anchor memory. Even brief exposure to framed headlines can exert lasting influence on user behavior.

\begin{figure*}
    \centering
    \includegraphics[width=0.99\linewidth]{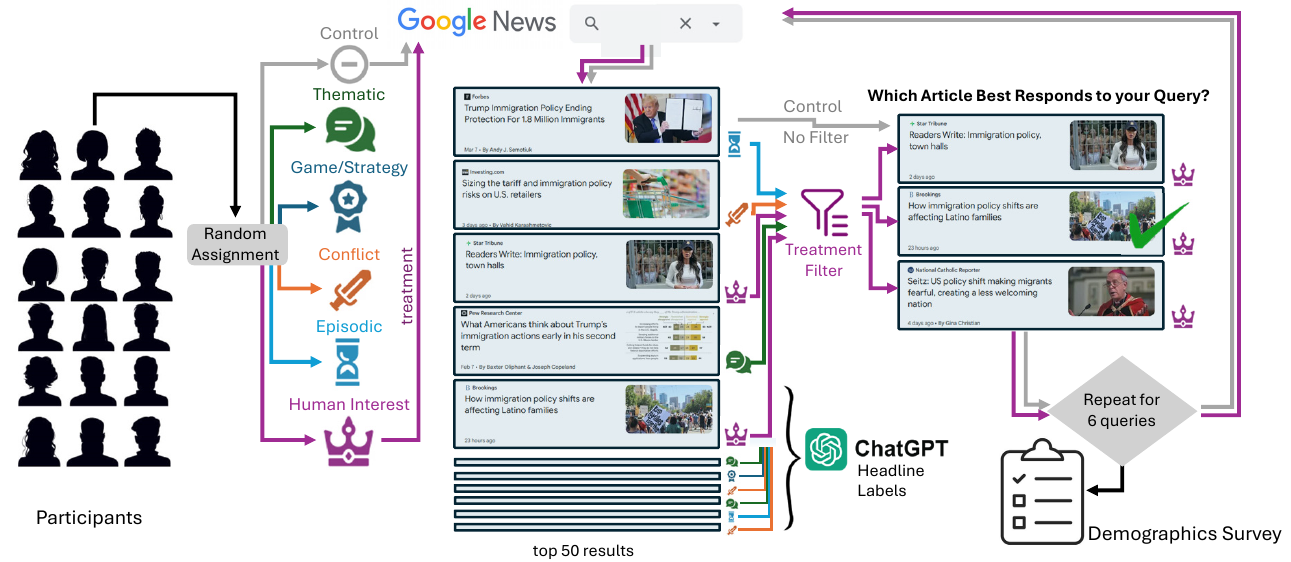}
    \caption{Experimental workflow illustrating the framing intervention and query reformulation loop. Participants were randomly assigned to one of six conditions (control or one of five framing treatments: conflict, game/strategy, human interest, episodic, or thematic). In each of six rounds, participants entered a news-related search query, viewed a subset of Google News headlines filtered according to their assigned frame, and selected one headline to read after a brief delay. We observed how exposure to specific frames influenced follow-up queries and their results. We measured changes in query content using semantic similarity, frame match rate, and linguistic characteristics such as concreteness and emotional valence.}
    \label{fig:method}
\end{figure*}

\subsection{Operationalizing Frames in Search Results}

We categorize headlines using six common frame types drawn from the political communication and media framing literature (see Table~\ref{tab:framing_categories}). This taxonomy provides a structured lens for investigating how headline framing influences user behavior across diverse news content~\cite{baden2019framing}.

While alternative framing taxonomies exist and may be more suitable for specific contexts~\cite{sullivan2023three}, the selected categories are among the most prevalent in news coverage~\cite{de2005news}. Headlines are not easily reducible to a single frame~\cite{semetko2000framing}, making frame selection especially important. The six frames we adopt are frequently observed in U.S. and European media and have been foundational in framing research~\cite{brants1998infotainment, dijk1988news, neuman1992common, lawrence2000game}.

\section{Experimental Design}

%To investigate how the framing of search results influences user behavior, w
We designed a between-subjects randomized controlled experiment (illustrated in Fig.~\ref{fig:method}) where participants were assigned to one of six experimental conditions, each corresponding to a specific news frame or a control group. The study examined how exposure to framed search results affected users' subsequent queries and selection behavior over multiple rounds of interaction. 

\paragraph{Procedure} 
Participants were randomly assigned to one of six frame conditions corresponding to journalistic frames: Game/Strategy, Conflict, Thematic, Episodic, Human Interest, and a control group. Each participant was asked to enter an initial news-related query into a custom web form that resembled a Google News search interface. The interface clearly indicated that it was part of an experiment and was not intended to deceive participants into believing it was a real search engine.

For each submitted query, the top 50 search results were retrieved in real time from Google News using the ScaleSERP API\footnote{\url{http://scaleserp.com}}. The API used U.S.-based proxies and returned English-language results; all other settings remained at their defaults.

Each headline was then automatically classified into one of the five frame types or labeled as other if no suitable frame was detected~\cite{poudel2025social}. Classification was performed using ChatGPT-4o Mini; the full classification prompt is included in the Appendix. Participants in the control condition were shown the top three headlines from the original search results, irrespective of frame. Participants in each treatment group were shown the top three headlines matching their assigned frame condition. If fewer than three matching headlines were available, the remaining results were randomly selected from the rest of the top 50 to ensure that three headlines were always displayed.

Participants were then asked to select one result to read, with a five-second delay before a selection could be made. This delay encouraged participants to review all options rather than clicking immediately. After making a selection, participants were prompted to enter a new query and repeat the process. Each participant completed six rounds in total, with the assigned frame condition held constant throughout. This design allowed us to observe how initial exposure to a particular frame shaped the trajectory of subsequent search behavior over time. We retrieved and classified over 157,000 headlines distributed as shown in Table~\ref{tab:framing_categories}.

\iffalse
% \ap{this is in 1 already}
\begin{table}[t]
    \centering
    \caption{Distribution in headline-frames from search results}
    \label{tab:headlinedistribution}
    \small{
    \begin{tabular}{ll}
    \toprule
     \textbf{Frame}  &  \textbf{Headlines} \\ \midrule
        Game/Strategy  &  42171 \\
        Conflict  &  28367 \\
        Thematic  &  56627 \\
        Episodic  &  10698 \\
        Human Interest  &  16593 \\
        (other)  &  3371 \\ \bottomrule
    \end{tabular}
    }
\end{table}

\fi

\paragraph{Outcome Measures} 
We recorded several aspects of participants’ information-seeking behavior throughout the experiment: (1) the sequence of search queries entered, (2) the search results presented in response to each query, and (3) the specific headlines selected by participants. These behavioral traces allow us to assess changes in search intent, framing exposure, and engagement across repeated interactions. At the conclusion of the task, participants completed a short demographic survey.

\paragraph{Participants}

We recruited $N=600$ participants from Prolific using the platform’s ``representative sample'' feature, which ensures demographic diversity aligned with U.S. census benchmarks. All participants were based in the United States, provided informed consent, and were compensated \$1.20 USD upon completion. The study was approved by the Institutional Review Board at the University of Notre Dame. 
% \textit{[redacted]}.

The sample included 47\% male and 51\% female participants (2\% preferred not to disclose). Participants resided in rural (15\%), suburban (53\%), and urban (32\%) areas. Educational attainment varied: 15\% had a high school diploma, 24\% had some college experience, 40\% held a college degree, and 22\% had a graduate degree. Racial/ethnic composition was 63\% White, 15\% African American, 10\% Latino, and 8\% Asian (4\% identified with other or multiple categories).

Participants also reported a range of political orientations: 12\% identified as very conservative, 24\% as somewhat conservative, 24\% as moderate, 24\% as somewhat liberal, and 16\% as very liberal. To assess baseline engagement with news search, participants were asked how frequently they used search engines to find news. Responses indicated that 11\% searched a few times per month or less, 25\% a few times per week, 17\% about once per day, and 45\% multiple times per day.

\section{Headlines Affect Follow-Up Queries}

Before turning to the effects of news framing, we begin with a foundational question: to what extent do search results shape the evolution of user queries, independent of their original intent? While prior work shows that users revise their queries based on both internal goals and external feedback~\cite{jansen2009patterns}, the degree to which retrieved content nudges users' thinking in new directions remains unclear.

\subsection{Query Evolution and Influence of Results}

To measure this influence, we embed queries and their retrieved headlines using a pre-trained Sentence Transformers model~\cite{reimers-2019-sentence-bert} and compute cosine similarity between: (1) a query and its immediate follow-up, and (2) the selected headline and the next query.

\begin{table}[t]
    \centering
    \caption{Factors Predicting How Users Reformulate Their Search Queries}
    \vspace{-0.3cm}
    \label{tab:ols_results}
    \small{
    \begin{tabular}{rr|c}
         \toprule 
         \textbf{Predictor} & \textbf{Effect} & \textbf{Query Influence} \\ \midrule 
         (Intercept) & –0.003$^{***}$ & \multirow{3}{*}{\begin{tikzpicture}
\begin{axis}[
    ybar,
    bar width=7pt,
    ymin=0,
    ymax=110,
    xmin=-0.5, xmax=1.5,
    ylabel=,
    xlabel={},
    xtick={-1},
    xticklabels={},
    nodes near coords,
    nodes near coords align={center},
    every node near coord/.append style={
        font=\tiny, anchor=center, yshift=4pt,
    },
    point meta=explicit symbolic, 
    yticklabel style={xshift=2pt},
    ytick={5,25,50,75,100},
    yticklabels={\tiny{0\%},{},\tiny{50\%},{},\tiny{100\%}},
    %nodes near coords,
    %nodes near coords align={vertical},
    width=3.3cm,
    height=2.5cm,
]
\addplot+[error bars/.cd,
    y dir=both,
    y explicit]
    coordinates {
    (0, 64.7) +- (0,2.0)[Prev Q.]
    (1, 35.3) +- (0,1.5)[Headline]
    };
\end{axis}
\end{tikzpicture}} \\
         Previous Query & 0.728$^{***}$ &   \\ 
         Retrieved Headlines & 0.398$^{***}$ &  \\ \bottomrule
         \multicolumn{3}{p{.93\linewidth}}{\scriptsize{*** indicates statistical significance at the 0.001 level. Higher values indicate stronger influence on the wording of the next query.}}
    \end{tabular}
    }
\end{table}

As shown in Table~\ref{tab:ols_results}, both the prior query and the retrieved headline significantly influence how users reformulate their next query. To interpret the relative magnitude of these effects, we compute the \textit{influence ratio} from the normalized effect sizes: approximately \textbf{65\%} of the semantic direction of the next query is attributable to the user’s own previous query, while \textbf{35\%} reflects the influence of the selected headline. This ratio highlights how even brief exposure to retrieved content can subtly steer user intent.

In other words, search results are not just a reflection of user goals; it is a site of influence, where platform-generated content feeds back into the query stream. This dynamic reveals a quiet but consequential loop: users shape their results, but results also shape users.

\paragraph{Quasi-Control Comparison} To test whether retrieved headlines causally influence query reformulation, we conducted a difference-in-differences analysis comparing real query–headline pairs to a quasi-control condition with randomly shuffled headlines. This preserved headline and query distributions while breaking their alignment. Follow-up queries in the real condition were significantly more semantically aligned with retrieved headlines than in the shuffled control (+0.036 cosine similarity; $p < .001$). This reinforces the interpretation that retrieved content, rather than internal goals alone, drives semantic shifts in query reformulation.

\paragraph{Decay Effects Across Query Turns} We next examined whether this influence persists over time. Semantic similarity between queries and prior headlines declined with each additional query round ($\beta = -0.0048$, $p < .001$), suggesting that framing effects are strongest immediately and fade with continued interaction.

\subsection{Summary}

This preliminary analysis supports the view that search results meaningfully guide the direction of user search behavior. While users' prior queries exert the strongest influence, retrieved headlines also shape future information-seeking, especially in the immediate next turn. This sets the stage for our core framing analysis by demonstrating that users are indeed responsive to the content of what they see.

\section{Framing Effects}

\subsection{Framing Influences Search Behavior}

Building on our preliminary finding that search results influence the trajectory of user queries, we now examine how specific types of framing shape user behavior. In particular, we hypothesize that \textbf{H1: Framing conditions increase the presence of their respective frames in subsequent queries compared to the control condition.}

Participants were randomly assigned to one of five frame-based treatment conditions (conflict, episodic, thematic, game/strategy, or human interest) or a control group with no frame filtering. Starting from their second query (\ie, after the initial query, which preceded treatment), we measured how participants responded to framed results over five rounds of search. %We now report results for each of the three framing hypotheses.

\subsubsection{Frames Influence Follow-up Queries}

We first tested whether concentrated exposure to framed headlines changes the follow-up queries. To do this, we computed the cosine similarity between the headline a user selected and their \textit{next} query. We then fit a linear mixed-effects model predicting similarity as a function of framing condition (relative to a control group) with random intercepts for each participant.

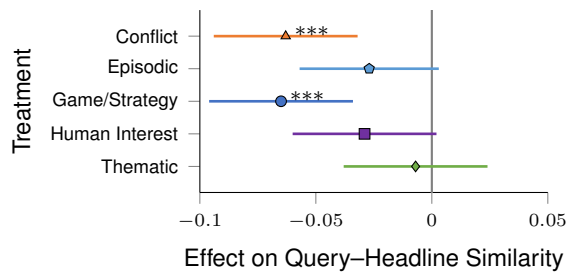
\begin{figure}
    \centering
    \begin{tikzpicture}
\begin{axis}[
    xticklabel style={
      /pgf/number format/fixed,
      /pgf/number format/precision=3
    },
    scaled x ticks=false,
    y dir=reverse,    
    xmin=-0.1,
    xmax=0.05,
    width=0.8\linewidth,
    height=4cm,
    axis lines*=left,
    ylabel={Treatment},
    xlabel={Effect on Query–Headline Similarity},
    yticklabels={Conflict, Episodic, Game/Strategy, Human Interest, Thematic},
    ytick={0,1,2,3,4},
    xtick={-0.1,-0.05,0,0.05},
    tick align=outside,
    enlarge y limits=0.2,
]
    % Plot the vertical line at x = 0 using an axis coordinate system
    \draw [gray, thick] ([yshift=-10pt]axis cs:0,5) -- 
                                 ([yshift=+10pt]axis cs:0,-2);

    % Just plot the dots and error bars (no connecting lines)
    \addplot+[
        only marks,
        mark=triangle*,
        mark options={black, fill=ppt-orange},
        mark size=2pt,
                error bars/.cd,
        x dir=both,
        x explicit,
        error mark=bar,
        error bar style={line width=1pt, ppt-orange},
    ]
    coordinates {
        (-0.063,0) +- (0.031,0)
    };

    \addplot+[
        only marks,
        mark=pentagon*,
        mark options={black, fill=ppt-blue},
        mark size=2pt,
        error bars/.cd,
        x dir=both,
        x explicit,
        error mark=bar,
        error bar style={line width=1pt, ppt-blue},
    ]
    coordinates {
        (-0.027,1) +- (0.030,0)
    };

    \addplot+[
        only marks,
        mark=*,
        mark options={black, fill=ppt-darkblue},
        mark size=2pt,
        error bars/.cd,
        x dir=both,
        x explicit,
        error mark=bar,
        error bar style={line width=1pt, ppt-darkblue},
    ]
    coordinates {
        (-0.065,2) +- (0.031,0)
    };

    \addplot+[
        only marks,
        mark=square*,
        mark options={black, fill=ppt-purple},
        mark size=2pt,
        error bars/.cd,
        x dir=both,
        x explicit,
        error mark=bar,
        error bar style={line width=1pt, ppt-purple},
    ]
    coordinates {
        (-0.029,3) +- (0.031,0)
    };

    \addplot+[
        only marks,
        mark=diamond*,
        mark options={black, fill=ppt-green},
        mark size=2pt,
        error bars/.cd,
        x dir=both,
        x explicit,
        error mark=bar,
        error bar style={line width=1pt, draw=ppt-green},
    ]
    coordinates {
        (-0.007,4) +- (0.031,0)
    };

    \node at (axis cs:-0.063,0) [xshift=10pt] {$^{***}$};
    \node at (axis cs:-0.065,2) [xshift=10pt] {$^{***}$};

\end{axis}
\end{tikzpicture}
    \caption{Estimated effects of each framing condition on query–headline similarity, relative to the control group. Points represent regression coefficients from the additive mixed-effects model. Horizontal lines show 95\% confidence intervals.}
    \label{fig:h1a_coeffplot}
\end{figure}

Figure~\ref{fig:h1a_coeffplot} shows the estimated effects of each frame condition on query-headline similarity. Participants exposed to \textit{Conflict} and \textit{Game/Strategy} frames submitted queries that were significantly less aligned with their selected headlines compared to those in the control group. The \textit{Human Interest} and \textit{Episodic} frames also showed negative trends, though their effects did not reach conventional levels of significance. The \textit{Thematic} condition was indistinguishable from control.

These results show some support for H1: certain frames, particularly those emphasizing conflict or strategic dynamics, disrupt alignment between what users read and how they continue searching. This may reflect how these frames shift attention toward emotionally or ideologically salient aspects of the news.

\subsubsection{Framing Shapes the Language of Follow-Up Queries}

Beyond semantic alignment, we also investigated whether framing influences the linguistic character of follow-up queries. Specifically, their level of concreteness and emotional content.

For concreteness, we computed a concreteness score for each query using a psycholinguistic dictionary that assigns word-level ratings based on imageability and specificity~\cite{brysbaert2014concreteness}. A one-way ANOVA revealed a marginal overall effect of frame condition on query concreteness ($F$(5, 2272) = 2.10, $p$ = .063). Post-hoc comparisons indicated that participants in the \textit{Episodic} condition wrote significantly more concrete queries than those in the \textit{Thematic} condition ($p$ = .037), consistent with the idea that episodic frames emphasize individual stories, while thematic frames promote broader, systemic interpretations.

For emotional content, we counted the number of emotional terms in each query using the NRC Emotion Lexicon~\cite{mohammad2013nrc}, focusing on words associated with anger, fear, or disgust. ANOVA again revealed a significant overall effect of frame condition ($F$(5, 2754) = 2.47, $p$ = .031), but no significant pairwise differences emerged in post-hoc tests. Contrary to our initial expectations, participants in the \textit{Conflict} condition did not produce more emotionally charged queries than those in other conditions.

Overall, these findings provide partial support for the hypothesis that framing shapes the linguistic style of follow up queries. Episodic frames increased concreteness, while emotional language was somewhat affected by frame type.

\subsection{Do Frames Persist Over Queries?}

Framing theory suggests that exposure to a particular narrative structure makes that frame more salient in subsequent interactions~\cite{entman1993framing, de2005news}. Based on these previous findings we hypothesize that \textbf{H2: Exposure to a specific frame in the search results will increase the likelihood that searches from later queries contain that same frame.}

\begin{figure}
    \centering
    \begin{tikzpicture}
\begin{axis}[
    xlabel={Marginal Prob. vs Control},
    xticklabel style={
      /pgf/number format/fixed,
      /pgf/number format/precision=3
    },
    scaled x ticks=false,
    y dir=reverse,
    width=0.85\linewidth,
    height=4.5cm,
    xmin=-0.1, xmax=0.1,
    axis lines*=left,
    enlarge y limits=0.2,
    legend pos=south east,
    yticklabels={Conflict, Episodic, Game/Strategy, Human Interest, Thematic},
    ytick={0,1,2,3,4},
    ylabel={Treatment}
]

% conflict
\addplot+[draw=ppt-orange, thick, line width=4pt, opacity=0.5 ]  coordinates {
    (-0.084422, 0)
    (0.05179, 0)
};

%control
\addplot+[ mark=star,mark options={black, fill=black}] coordinates {
    (-0.03788, 0) };
%human interest
\addplot+[ mark=square*,mark options={black, fill=ppt-purple}] coordinates {
    (-0.017106, 0) };
%thematic
\addplot+[ mark=diamond*,mark options={black, fill=ppt-green}] coordinates {
    (-0.084422, 0) };
%conflict    
\addplot+[ mark=triangle*,mark options={black, fill=ppt-orange}] coordinates {
    (0.051794, 0) };
%game
\addplot+[ mark=*,mark options={black, fill=ppt-darkblue}] coordinates {
    (-0.057539, 0) };
%episodic
\addplot+[ mark=pentagon*,mark options={black, fill=ppt-blue}] coordinates {
    (0.009165, 0) };

%conflict    
\addplot+[ mark=triangle*,mark options={black, fill=ppt-orange}] coordinates {
    (-0.1, 0) };    

\node at (axis cs:0.051794,0) [xshift=4pt] {$^{*}$};

%episodic
\addplot+[draw=ppt-blue, thick, line width=4pt, opacity=0.5 ]  coordinates {
    (-0.025436, 1)
    (0.023033, 1)
};

%control
\addplot+[ mark=star,mark options={black, fill=black}] coordinates {
    (0.020251, 1) };
%human interest
\addplot+[ mark=square*,mark options={ppt-blue, fill=ppt-purple}] coordinates {
    (-0.025436, 1) };
%thematic
\addplot+[ mark=diamond*,mark options={ppt-blue, fill=ppt-green}] coordinates {
    (-0.014635, 1) };
%conflict    
\addplot+[ mark=triangle*,mark options={ppt-blue, fill=ppt-orange}] coordinates {
    (-0.014406, 1) };
%game
\addplot+[ mark=*,mark options={ppt-blue, fill=ppt-darkblue}] coordinates {
    (0.023033, 1) };
%episodic
\addplot+[ mark=pentagon*,mark options={ppt-blue, fill=ppt-blue}] coordinates {
    (0.012906, 1) };

\addplot+[ mark=pentagon*,mark options={black, fill=ppt-blue}] coordinates {
    (-0.1, 1) };

%game
\addplot+[draw=ppt-darkblue, thick, line width=4pt, opacity=0.7 ] coordinates {
    (-0.047896, 2)
    (0.048858, 2)
};

%control
\addplot+[ mark=star,mark options={black, fill=black}] coordinates {
    (-0.018359, 2) };
%human interest
\addplot+[ mark=square*,mark options={black, fill=ppt-purple}] coordinates {
    (0.006274, 2) };
%thematic
\addplot+[ mark=diamond*,mark options={black, fill=ppt-green}] coordinates {
    (0.048858, 2) };
%conflict    
\addplot+[ mark=triangle*,mark options={black, fill=ppt-orange}] coordinates {
    (-0.006932, 2) };
%game
\addplot+[ mark=*,mark options={black, fill=ppt-darkblue}] coordinates {
    (-0.010925, 2) };
%episodic
\addplot+[ mark=pentagon*,mark options={black, fill=ppt-blue}] coordinates {
    (-0.047896, 2) };

\addplot+[ mark=*,mark options={black, fill=ppt-darkblue}] coordinates {
    (-0.1, 2) };

%human
\addplot+[draw=ppt-purple, thick, line width=4pt, opacity=0.5 ] coordinates {
    (-0.005278, 3)
    (0.031084, 3)
};

%control
\addplot+[ mark=star,mark options={black, fill=black}] coordinates {
    (0.031084, 3) };
%human interest
\addplot+[ mark=square*,mark options={black, fill=ppt-purple}] coordinates {
    (0.023151, 3) };
%thematic
\addplot+[ mark=diamond*,mark options={black, fill=ppt-green}] coordinates {
    (0.010313, 3) };
%conflict    
\addplot+[ mark=triangle*,mark options={black, fill=ppt-orange}] coordinates {
    (-0.005278, 3) };
%game
\addplot+[ mark=*,mark options={black, fill=ppt-darkblue}] coordinates {
    (0.000278, 3) };    
%episodic
\addplot+[ mark=pentagon*,mark options={black, fill=ppt-blue}] coordinates {
    (0.000289, 3) };

\addplot+[ mark=square*,mark options={black, fill=ppt-purple}] coordinates {
    (-0.1, 3) };

%theme
\addplot+[draw=ppt-green, thick, line width=4pt, opacity=0.5 ] coordinates {
    (-0.024988, 4)
    (0.054303, 4)
};

%control
\addplot+[ mark=star,mark options={black, fill=black}] coordinates {
    (-0.000425, 4) };
%human interest
\addplot+[ mark=square*,mark options={black, fill=ppt-purple}] coordinates {
    (0.006616, 4) };
%thematic
\addplot+[ mark=diamond*,mark options={black, fill=ppt-green}] coordinates {
    (0.038655, 4) };
%conflict    
\addplot+[ mark=triangle*,mark options={black, fill=ppt-orange}] coordinates {
    (-0.024988, 4) };
%game
\addplot+[ mark=*,mark options={black, fill=ppt-darkblue}] coordinates {
    (0.045651, 4) };
%episodic
\addplot+[ mark=pentagon*,mark options={black, fill=ppt-blue}] coordinates {
    (0.054303, 4) };

\addplot+[ mark=diamond*,mark options={black, fill=ppt-green}] coordinates {
    (-0.1, 4) };

    % Plot the vertical line at x = 0 using an axis coordinate system
\draw [gray, thick] ([yshift=-10pt]axis cs:0,-2) -- 
                                 ([yshift=+10pt]axis cs:0,6);
\end{axis}
\end{tikzpicture}
    \caption{Difference-in-differences estimates showing how much each frame-condition pair deviates from the control group in follow-up query framing. Positive values indicate greater increases in frame usage relative to control. * indicates statistical significance at the 0.05 level.}
    \label{fig:did_plot}
\end{figure}
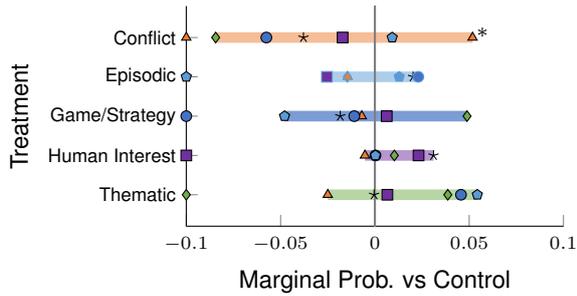

Because user queries are typically short and ambiguous, we do not attempt to classify the frame of each query directly. Instead, we infer query framing indirectly by examining the distribution of frames in the retrieved results from subsequent queries. If exposure to a particular frame influences user behavior, we expect users to submit follow-up queries that elicit headlines framed in a similar way.

To assess whether observed changes in query behavior are attributable to exposure to framed search results, rather than natural variation or individual intent, we use a \textit{difference-in-differences} (DiD) approach. 

In our case, DiD compares the change in frame usage between the initial query (Q1) and the next query (Q2) (\ie, post-treatment) for users exposed to a given frame condition compared to the change observed in a control group. Formally, the DiD estimate for a given frame and condition is defined as:
\begin{align*}
\text{DiD} &= \left( \Delta_\text{treat} \right) - \left( \Delta_\text{cntrl} \right) \\
&= \left( \text{Q2}_{\text{treat}} - \text{Q1}_{\text{treat}} \right)
- \left( \text{Q2}_{\text{cntrl}} - \text{Q1}_{\text{cntrl}} \right)
\end{align*}

% \[
% \text{DiD} = (\text{Q2}_{\text{treat}} - \text{Q1}_{\text{treat}}) - (\text{Q2}_{\text{cntrl}} - \text{Q1}_{\text{cntrl}})
% \]

This expression captures how much more (or less) a given frame increased under treatment relative to its natural trajectory in the absence of framing intervention.

To control for baseline variation in frame preferences, we apply this comparison \textit{within matched frame categories} (\eg, measuring how use of the \textit{Conflict} frame evolves in the conflict condition versus the control condition). The control group provides a reference for expected change due to organic search behavior, while the treatment group reflects potential shifts driven by exposure to framed headlines.

We compute the mean change in frame proportions for each participant and compare the treatment and control groups using two-sample $t$-tests. To adjust for multiple comparisons across frame-condition pairs, we consider Bonferroni-adjusted $p$-values.

\paragraph{Conflict Begets Conflict}
As shown in Figure~\ref{fig:did_plot}, the only condition to show a notable increase relative to control is the \textit{conflict} frame in the conflict condition, which yielded a large mean difference (+8.9\%) and a nominally significant $p$-value ($p = 0.015$). This finding reinforces the idea that conflict primes further conflict and aligns with prior research suggesting that conflict frames increase perceived stakes, activate adversarial schemas, and persist longer in cognitive processing~\cite{shen2004effects}.

Otherwise, we find limited evidence that headline framing systematically alters the use of corresponding frames in follow-up queries. Other frame--condition pairs show smaller or inconsistent changes, suggesting that while framing can shape short-term attention and interpretation, its influence on query framing is subtle.

\subsection{Do Frames Accumulate?}

\textbf{H3: Queries following exposure to a particular frame will increasingly reflect that frame over time.} This hypothesis draws on theories of cognitive carryover and priming: exposure to a specific narrative frame may lead users to unconsciously persist with that framing in subsequent queries. If framing effects accumulate, we would expect users to retrieve increasingly frame-aligned results as their session progresses.

To test this, we calculated the proportion of retrieved headlines that matched each participant’s assigned frame across all six query rounds. As in the prior analysis, we compared these frame match rates against a baseline from the control group. However, instead of a simple before/after comparison, we modeled change over time using a mixed-effects linear regression, with frame condition and query number as predictors. This design allows us to test whether frame alignment increases progressively over the course of a search session, \ie, whether conflict primes more conflict, or episodic frames lead to increasingly concrete queries.

The results (not illustrated) show no evidence of accumulation. Although some frame conditions initially yielded higher match rates than control, there was no significant interaction between condition and query number. In other words, participants did not increasingly align with their assigned frame as the session progressed.

\section{Exploratory Analysis}

\paragraph{E1: Are Frequent News Seekers More Responsive to Framing?}

To test whether frequent search users are more sensitive to framing, we included a self-reported news search frequency in our model as a moderator. The interaction between frame condition and search frequency was positive but not uniformly significant across all frames. Participants who reported using search engines more frequently to find news exhibited modestly higher frame match rates following exposure, suggesting a greater susceptibility to framing effects. But the effect did not persist after multiple comparison correction and should be interpreted cautiously.

\begin{figure}
    \centering
    \begin{tikzpicture}
\begin{axis}[
legend pos=north west,
xticklabel style={rotate=45, anchor=south east, xshift=10pt, yshift=-10pt},
width=.98\linewidth,
height=5cm,
xticklabels= {Conflict,Episodic,Game/Strategy,Human Interest,Thematic},
ybar=2pt,
bar width=0.15,
xtick={0,1,2,3,4},
ymin=-0.05,ymax=0.06,
ylabel={Marginal Prob.},
xlabel={Returned Headline Frames},
legend style={fill=none},
legend columns=3,
    yticklabel style={
      /pgf/number format/fixed,
      /pgf/number format/precision=3
    },
    scaled y ticks=false,
]

%liberal
\addplot+[fill=ppt-darkblue, fill opacity=.5, ppt-darkblue, error bars/.cd,
y dir=both,y explicit, error bar style={black}]
coordinates {
    (0, 0.006673) +- (0, 0.013)
    (1, -0.01527) +- (0, 0.006)
    (2, 0.004429) +- (0, 0.009)
    (3, -0.007388) +- (0, 0.010)
    (4, 0.012879) +- (0, 0.013)
    %(other, 0.021) +- (0, 0.002)
    };

%moderate
\addplot+[fill=ppt-purple, fill opacity=.5, ppt-purple, error bars/.cd,
y dir=both,y explicit, error bar style={black}]
coordinates {
    (0, -0.033464) +- (0, 0.006)
    (1, -0.001164) +- (0, 0.006)
    (2, 0.025342) +- (0, 0.009)
    (3, -0.009448) +- (0, 0.010)
    (4, 0.020477) +- (0, 0.016)
    %(other, 0.021) +- (0, 0.002)
    };

%conservative
\addplot+[fill=ppt-red, fill opacity=.5, ppt-red, error bars/.cd,
y dir=both,y explicit, error bar style={black}]
coordinates {
    (0, -0.020193) +- (0, 0.012)
    (1, 0.003981) +- (0, 0.006)
    (2, 0.010632) +- (0, 0.013)
    (3, -0.009049) +- (0, 0.006)
    (4, 0.016894) +- (0, 0.016)
    %(other, 0.017) +- (0, 0.002)
    };

% Significance bracket over x=0 (conflict)
\draw[thick] (axis cs:-0.2,0.025) -- (axis cs:0.0,0.025); % horizontal line
\draw[thick] (axis cs:-0.2,0.025) -- (axis cs:-0.2,0.02); % left vertical tick
\draw[thick] (axis cs:0.0,0.025) -- (axis cs:0.0,0.02); % right vertical tick
\node at (axis cs:-0.1,0.03) {\textasteriskcentered}; % significance mark

% Significance bracket over x=0 (conflict)
\draw[thick] (axis cs:-0.2,0.035) -- (axis cs:0.2,0.035); % horizontal line
\draw[thick] (axis cs:-0.2,0.035) -- (axis cs:-0.2,0.03); % left vertical tick
\draw[thick] (axis cs:0.2,0.035) -- (axis cs:0.2,0.03); % right vertical tick
\node at (axis cs:0,0.04) {\textasteriskcentered}; % significance mark

\draw [gray, thick] ([xshift=-10pt]axis cs:-1,0) -- 
                                 ([xshift=+10pt]axis cs:6,0);

\legend{Liberal, Moderate, Conservative}
\end{axis}

\end{tikzpicture}
    \caption{Marginal change in frame usage across political identity groups, relative to the control condition. Liberal participants exposed to conflict-framed headlines exhibit a significantly greater increase in conflict framing than both moderates and conservatives%, suggesting heightened responsiveness to ideologically resonant frames. 
    Error bars represent 95\% confidence intervals.}
    \label{fig:politics}
\end{figure}
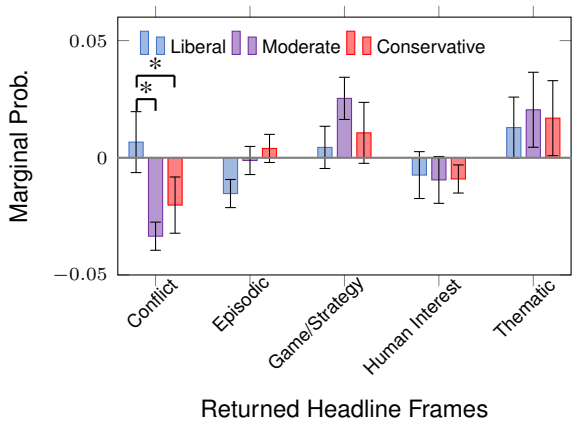

\paragraph{E2: Does Political Orientation Moderate Framing Effects?}
To test whether political identity shapes responsiveness to framing, we examined marginal changes in frame use relative to the control group across ideological groups. As shown in Figure~\ref{fig:politics}, the strongest effect emerged for the conflict frame: liberal participants exposed to conflict-framed headlines exhibited significantly greater increases in conflict framing in their follow-up queries compared to moderates and conservatives. This pattern suggests that frame-consistent interpretations may resonate more strongly with participants whose beliefs align with the emotional or ideological tone of the frame, offering support for E2 and broader theories of motivated reasoning.

\section{Discussion}

This study shows that subtle variations in headline framing significantly influence user behavior in search environments. Even a single exposure can alter how users reformulate queries and what narrative structures persist in follow-up results, revealing a dynamic interplay between algorithmic curation and cognitive bias.

We find strong support for H1: framing shapes immediate search behavior. Conflict and strategy frames reduce semantic alignment with headlines and lead to emotionally or ideologically salient query reformulations. H2 receives partial support: exposure to certain frames increases the likelihood of retrieving similarly framed results in the next turn, but these effects diminish quickly, offering no support for H3. Framing shapes intent, but its influence fades across rounds.

Exploratory findings highlight that individual traits moderate framing effects. Frequent news searchers (E1) are more responsive to frame cues, and political orientation (E2) amplifies effects of conflict framing—particularly among liberal users—suggesting that ideologically congruent frames may more strongly guide interpretation.

Overall, these findings underscore that search engines do more than reflect user intent; they actively shape it. Even small changes in framing can nudge users along different information paths, influencing not just what they see, but what they search for next.

\section{Conclusions}

Framing effects in search results are subtle but consequential. This study demonstrates that the narrative structure of headlines can influence the trajectory of user queries, reinforcing certain interpretations and shaping short-term information seeking. While these effects attenuate over time, their immediate impact underscores the importance of framing as a mechanism of cognitive and algorithmic influence. Understanding this dynamic is essential for designing search systems that support informed, reflective engagement with news content.

\section{Limitations}

Although our findings provide evidence that headline framing influences search behavior, several limitations remain.

First, the study was conducted in a controlled experimental setting. Although this design allows us to isolate the effects of framing, it may not fully capture the complexity of real-world search behavior, where goals are more varied, distractions are greater, and attention is more fragmented.

Second, although we model user interaction across multiple query rounds, the total session length is only six queries---a few minutes in duration. It is possible that framing effects, especially those related to accumulation (H3), may require longer exposure periods or more organic engagement to manifest more strongly.

Third, our operationalization of framing relies on a discrete set of frame categories drawn from political communication literature. While these are well-established and cover common types such as conflict, thematic, and human interest, they may not capture all the nuanced or hybrid frames that users interpret in context. Additionally, frame classification may carry some subjectivity.

Fourth, our sample is drawn from an online participant pool and may not be representative of the broader population. In particular, exploratory findings related to political identity (E2) should be interpreted cautiously given potential imbalances in ideological self-identification and the correlational nature of these subgroup comparisons.

Finally, while we observed statistically significant effects, many were small in magnitude. This highlights both the subtlety of framing influence and the need for large samples or complementary measures (\eg response time, gaze tracking, or qualitative follow-ups) to fully understand the mechanisms at work.

Future work should explore how these framing effects unfold in more naturalistic settings, over longer sessions, or across different types of platforms. Additional attention should also be paid to how repeated exposure, emotional tone, and algorithmic feedback loops interact with individual predispositions. By treating search behavior as both a cognitive and sociotechnical process, we can better understand the forces that govern attention, interpretation, and ultimately, belief.

\section{Acknowledgement}
We would like to thank Yifan Ding and Matthew Facciani for their helpful discussions.

\appendix

\bibliography{ref}
\bibliographystyle{acl_natbib}

\clearpage
\newpage

\clearpage
\appendix
\onecolumn
\section*{APPENDIX}
\label{appendix}
\addcontentsline{toc}{section}{APPENDIX}
\setcounter{table}{0}
\renewcommand{\thetable}{A\arabic{table}} 
\newcounter{myboxcounter}{}
\setcounter{myboxcounter}{0}  % Reset the counter for mybox

\newtcolorbox{mybox}[2][]{%
width=\textwidth,
  colframe=black,
  colback=white,
  sharp corners=south,
  title=Box~\refstepcounter{myboxcounter}\themyboxcounter: #2,
  #1
}

\section{Framing Classification Prompt}

\begin{mybox}{Framing Classification Prompt}
\label{box:user_behavior}
You are an assistant trained to classify news headlines into one of the following generic frames based on their dominant focus. Below are the definitions of each frame: \\

    1. conflict\\
    - Description: Presents events as a conflict between competing actors, issues, or interpretations.\\
    2. game\_strategy\\
    - Description: Focuses on the efforts of actors to gain support, influence, or achieve specific goals.\\
    3. thematic\\
    - Description: Centers on the substantive content of public concerns and issues. \\
    4. human\_interest\\
    - Description: Narrates events from the perspective of individuals affected by the issues or events.\\
    5. episodic\\
    - Description: Presents specific events or episodes without extensive context or connection to broader themes.\\
    6. other\\
    - Description: if the news does not fit into any of the above frames\\
    
    Task:\\
    Given a news headline, classify it into one of the above frames by selecting the most appropriate single frame that best represents the headline's primary focus. Respond only with the name of the frame (e.g., "conflict"). \\
    
    Additional Notes:\\
    - Select Only One Frame: Assign only the dominant frame that best fits the headline, even if multiple frames seem relevant.\\
    - Consistency: Use the exact frame names provided in the definitions for clarity and consistency.\\
    - Clarity: Ensure that the classification is based solely on the headline's content without requiring external context.\\
    Example:\\
    - Headline: "Local Hero Rescues Family from Burning Building"\\
    - Frame: Human Interest Frame\\

    Now, classify the following headline: \\
    \textbf{Headline:} \textit{title}
    
\end{mybox}

\vspace{0.3cm}

\end{document}